\documentclass{article} 
\usepackage{iclr2019,times}
\usepackage{amssymb}
\usepackage{float}
\usepackage{hyperref}
\usepackage{url}
\usepackage{bm}
\usepackage{graphicx,array} \graphicspath{{figures/}}
\usepackage{amsmath}
\usepackage{color}
\usepackage{caption}
\usepackage{subcaption}

\newcommand{\figref}[1]{figure~\ref{fig:#1}}
\newcommand{\Figref}[1]{Figure~\ref{fig:#1}}

\renewcommand{\b}[1]{{\bm{#1}}}   

\renewcommand{\L}{\b{L}}

\newcommand{\U}{\b{U}}

\newcommand{\trans}{^\intercal}

\newcommand{\bLambda}{\b{\Lambda}}

\newcommand{\bO}{\mathcal{O}}

\title{DeepSphere: towards an equivariant graph-based spherical CNN}

\author{Michaël Defferrard \\
Institute of Electrical Engineering \\
EPFL, Lausanne, Switzerland \\
\texttt{michael.defferrard@epfl.ch}
\And
Nathanaël Perraudin \\
Swiss Data Science Center (SDSC) \\
Zurich, Switzerland \\
\texttt{nathanael.perraudin@sdsc.ethz.ch}
\And
Tomasz Kacprzak \& Raphael Sgier \\
Institute for Particle Physics and Astrophysics, ETH Zurich, Switzerland \\
\texttt{\{tomaszk,rsgier\}@phys.ethz.ch} \\
}

\iclrfinalcopy 
\begin{document}

\maketitle

\begin{abstract}
	Spherical data is found in many applications.
	By modeling the discretized sphere as a graph, we can accommodate non-uniformly distributed, partial, and changing samplings.
	Moreover, graph convolutions are computationally more efficient than spherical convolutions.
	As equivariance is desired to exploit rotational symmetries, we discuss how to approach rotation equivariance using the graph neural network introduced in \citet{defferrard2016convolutional}.
	Experiments show good performance on rotation-invariant learning problems.
	Code and examples are available at \url{https://github.com/SwissDataScienceCenter/DeepSphere}.
\end{abstract}

\begin{figure}[h]
	\includegraphics[height=0.25\linewidth]{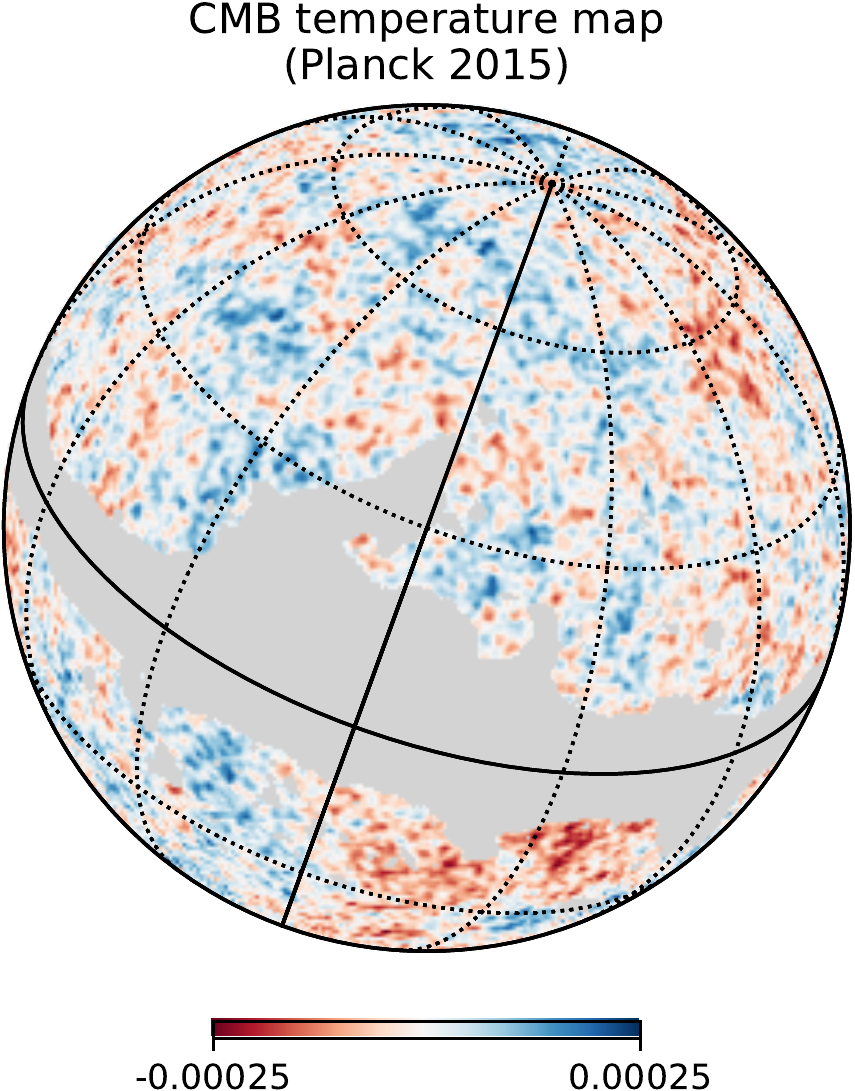}
	\hfill
	\includegraphics[height=0.25\linewidth]{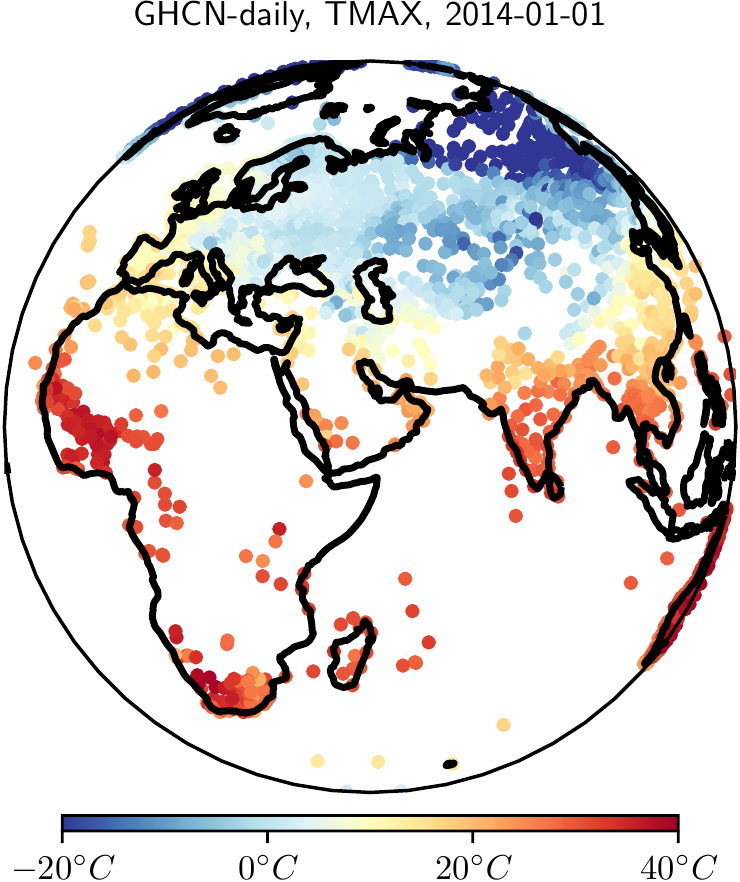}
	\hfill
	\includegraphics[height=0.25\linewidth]{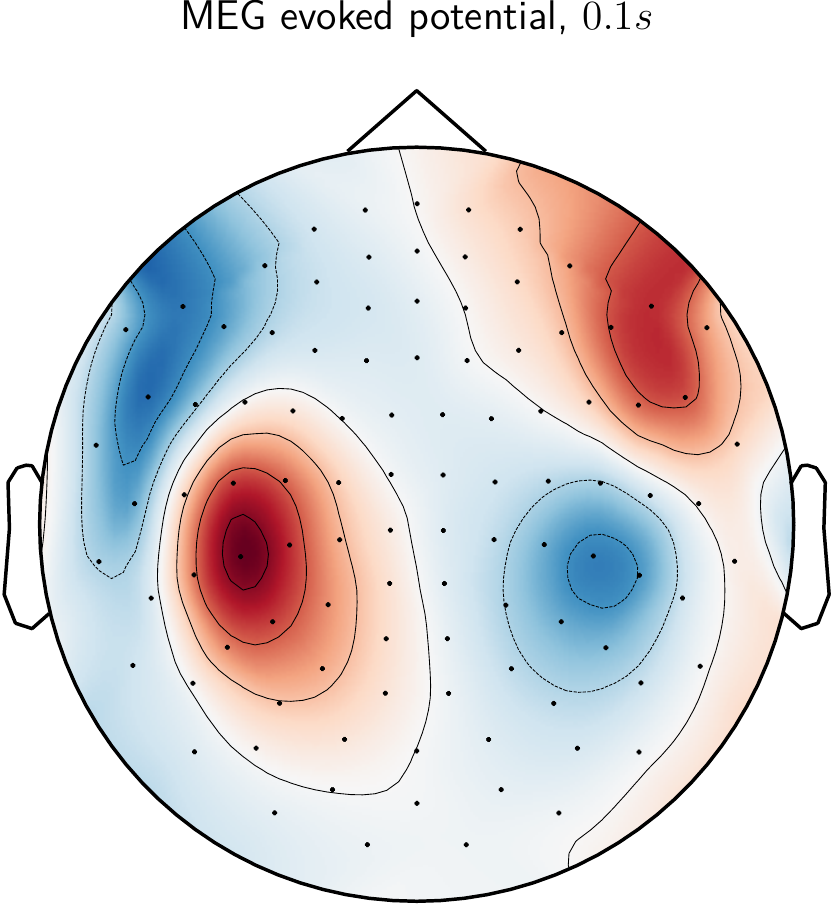}
	\hfill
	\includegraphics[height=0.25\linewidth]{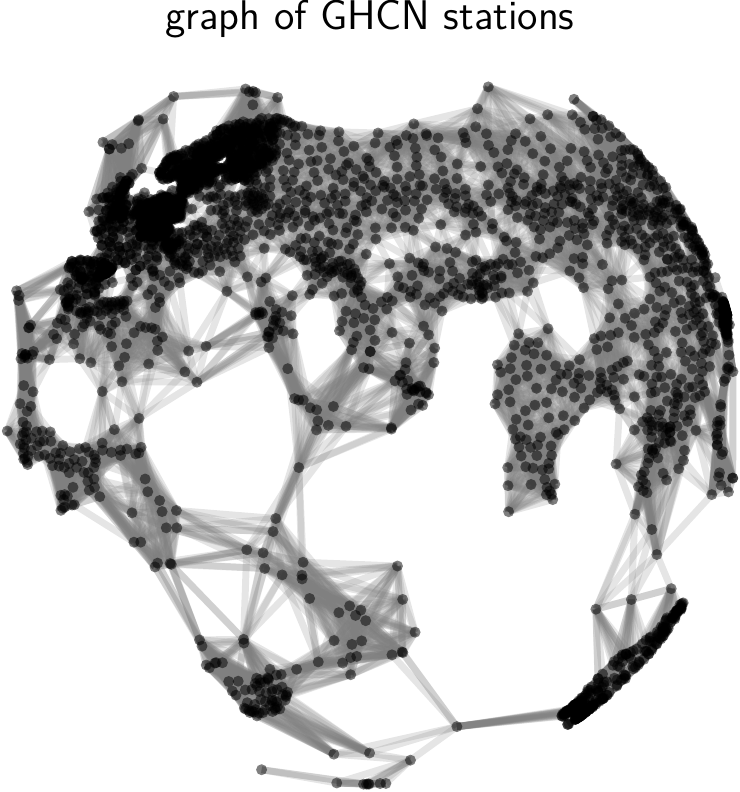}
	\caption{
		Examples of intrinsically spherical data: (left) the cosmic microwave background (CMB) temperature from \citet{planck2015overview},
		(middle) daily maximum temperature from the Global Historical Climatology Network (GHCN),\protect\footnotemark (right) brain activity recorded through magnetoencephalography (MEG).\protect\footnotemark
		For those examples, a rigid full-sphere pixelization is not ideal: the Milky Way's galactic plane masks observations, brain activity is measured on the scalp only, and the position and density of weather stations is arbitrary and changes over time.
		Graphs can faithfully and efficiently represent sampled spherical data by placing vertices where data has been measured.
	}
	\label{fig:examples}
\end{figure}
\footnotetext{\scriptsize\url{https://www.ncdc.noaa.gov/ghcn-daily-description}}
\footnotetext{\scriptsize\url{https://martinos.org/mne/stable/auto_tutorials/plot_visualize_evoked.html}}

\section{Introduction}




Graphs have long been used as models for discretized manifolds: for example to smooth meshes \citep{taubin1996meshsmoothing}, reduce dimensionality \citep{belkin2003laplacian}, and, more recently, to process signals \citep{shuman2013gsp}.
Along
Euclidean spaces, the sphere is one of the most commonly encountered manifold: it is notably used to represent omnidirectional images, global planetary data (in meteorology, climatology, geology, geophysics, etc.), cosmological observations, and brain activity measured on the scalp (see \figref{examples}).
Spherical convolutional neural networks (CNNs) have been developed to work with some of these modalities \citep{cohen2018sphericalcnn, perraudin2018deepsphere, khasanova2017graphomni, su2017sphericalcnn, coors2018sphericalcnn,
jiang2019sphericalcnn}.

Spherical data can be seen as a continuous function
that is sampled at discrete locations.
As it is impossible to construct a regular discretization of the sphere, there is no perfect spherical sampling.
Schemes have been engineered for particular applications and come with trade-offs \citep{gorski2005healpix,glesp}.
However, while sampling locations can be precisely controlled in some cases (like the CMOS sensors of an omni-directional camera), sensors might in general be non-uniformly distributed, cover only part of the sphere, and move (see \figref{examples}).
Modeling the sampled sphere as a discrete graph has the potential to faithfully and efficiently represent sampled spherical data by placing vertices where data has been measured: no need to handle missing data or to interpolate to some predefined sampling, and no waste of memory or precision due to over- or under-sampling.
Graph-based spherical CNNs have been proposed in \citet{khasanova2017graphomni} and \citet{perraudin2018deepsphere}.
Moreover, graph convolutions have a low computational complexity of $\bO(N_{pix})$, where $N_{pix}$ is the number of considered pixels.
Methods based on proper spherical convolutions, like \citet{cohen2018sphericalcnn} and \citet{esteves2017sphericalcnn}, cost $\bO(N_{pix}^{2/3})$ operations.




Finally, like classical 2D CNNs are equivariant to translations, we want spherical CNNs to be equivariant to 3D rotations \citep{cohen2016equivariance, kondor2018equivariance}.
A rotation-equivariant CNN detects patterns regardless of how they are rotated on the sphere: it exploits the rotational symmetries of the data through weight sharing.
Realizing that, spheres can be used to support data which does not intrinsically live on a sphere but have rotational symmetries \citep[for 3D objects and molecules]{cohen2018sphericalcnn, esteves2017sphericalcnn}.
In this contribution we present DeepSphere~\citep{perraudin2018deepsphere}, a spherical neural network leveraging graph convolution for its speed and flexibility.
Furthermore, we discuss the rotation equivariance of graph convolution on the sphere.



\begin{figure}[t]
	\centering
	\includegraphics[width=0.9\linewidth]{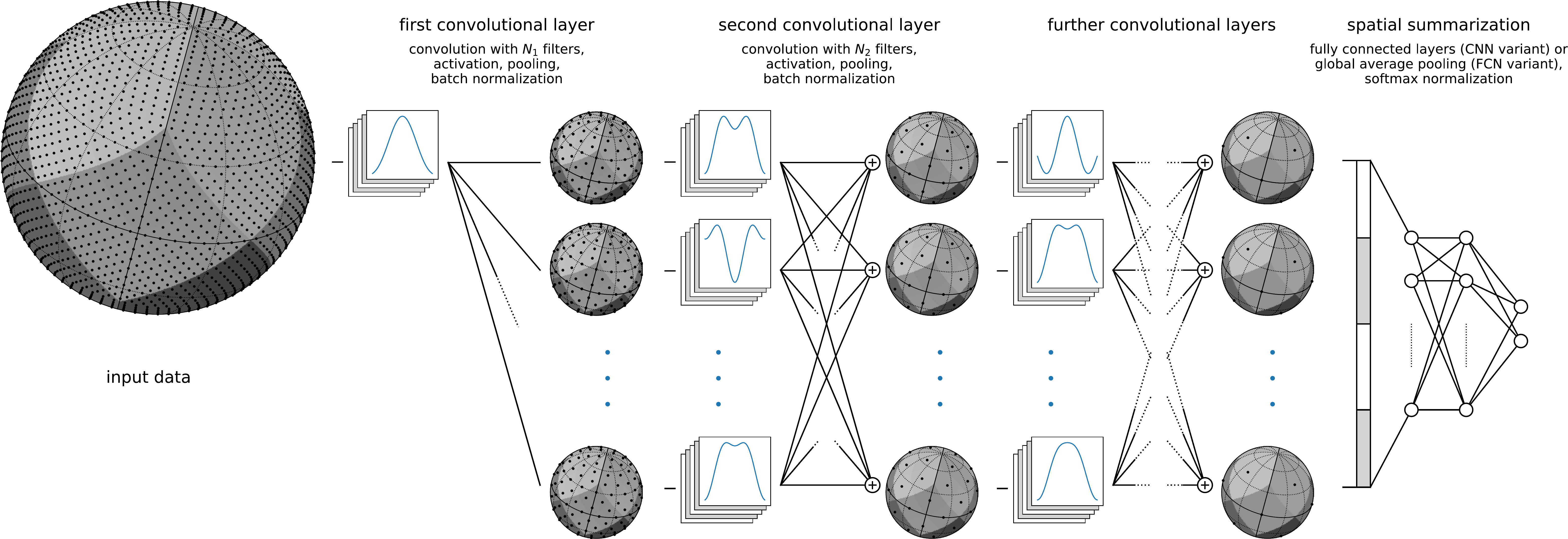}
	\caption{Example architecture.
		Global tasks need a spatial summarization: the FCN variant is rotation invariant (and accepts inputs of varying sizes), while the CNN variant is not.
		Dense tasks (when the output lives on the sphere, like segmentation) are rotation equivariant.
	}
	\label{fig:architecture}
\end{figure}

\section{Method}

Our method relies on the graph signal processing framework~\citep{shuman2013gsp}, which highly relies on the spectral properties of the graph Laplacian operator.
In particular, the Fourier transform is defined as the projection of the signal on the eigenvectors of the Laplacian, and the graph convolution as a multiplication in the Fourier domain.
It turns out that the graph convolution can be accelerated by being performed directly in the vertex domain
\citep{hammond2011wavelets}.

DeepSphere leverages graph convolutions
to achieve the following properties: (i) computational efficiency, (ii) adaptation to irregular sampling, and (iii) close to rotation equivariance.
An example architecture is shown in \figref{architecture}.
The main idea is to model the discretised sphere as a graph of connected pixels: the length of the shortest path between two pixels is an approximation of the geodesic distance between them.
We use the graph CNN formulation introduced in \citep{defferrard2016convolutional}, and a pooling strategy that exploits a hierarchical pixelisation of the sphere to analyse the data at multiple scales.
The current implementation of DeepSphere relies on the Hierarchical Equal Area isoLatitude Pixelisation (HEALPix)~\citep{gorski2005healpix}, a popular sampling used in cosmology and astrophysics.
See \citet{perraudin2018deepsphere} for details.
DeepSphere is, however, easily used with other samplings as only two elements depend on it: (i) the choice of neighboring pixels when building the graph, and (ii) the choice of parent pixels when building the hierarchy.

The flexibility of modeling the data domain with a graph allows one to easily model data that spans only a part of the sphere, or data that is not uniformly sampled.
Furthermore, using a $k$-nearest neighbors graph, the convolution operation costs $\mathcal{O}(N_{pix})$ operations.
This is the lowest possible complexity for a convolution without approximations.
Nevertheless, while the graph framework offers great flexibility, its ability to faithfully represent the underlying sphere highly depends on the sampling locations and the graph construction.
This should not be neglected since the better the graph represents the sphere, the closer to rotation equivariant the graph convolution will be.





\section{Harmonics and equivariance}



\begin{figure}
\begin{minipage}[c]{.21\linewidth}
	\centering
	\includegraphics[width=\linewidth]{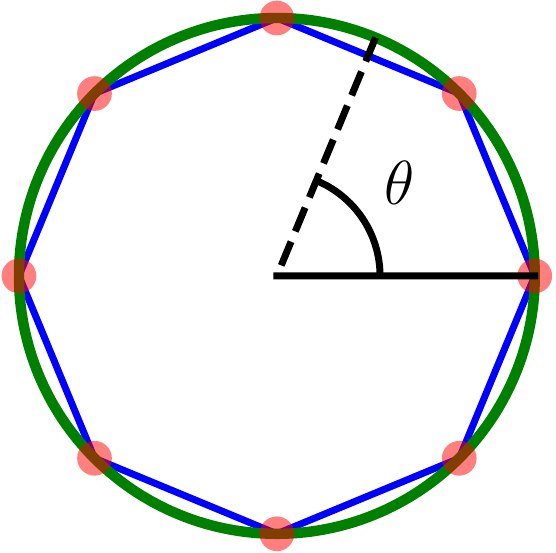}
	\caption{circle $\mathcal{C}^1$ (green), regular samples and graph vertices (red), graph edges (blue).}
	\label{fig:ring}
\end{minipage}
\hfill
\begin{minipage}[c]{.7\linewidth}
	\centering
	\includegraphics[width=\linewidth]{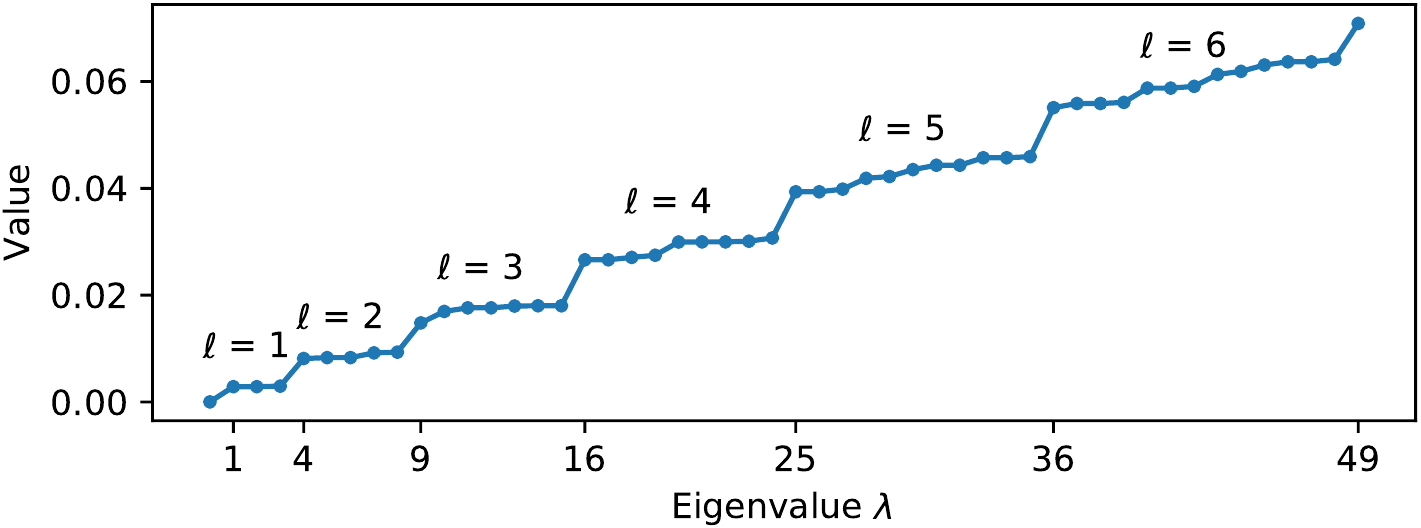}
	\caption{The eigenvalues $\bLambda$ of the graph Laplacian $\L = \U \bLambda \U\trans$ are clearly organized in groups. Each group corresponds to a degree $\ell$ of the spherical harmonics. Each degree has $2\ell + 1$ orders.
	}
	\label{fig:graph_eigenvalues}
\end{minipage}
\end{figure}

\begin{figure}
	\centering
	\includegraphics[trim={0 6.37cm 0 0},clip,height=0.15\linewidth]{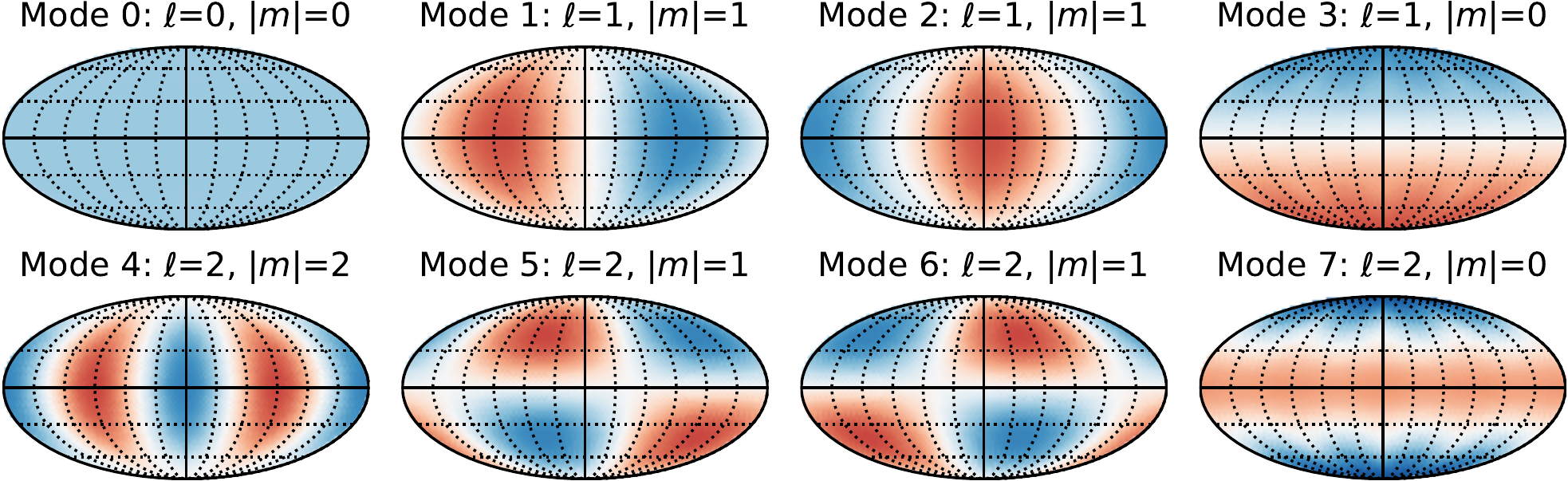}
	\hfill
	\includegraphics[height=0.15\linewidth]{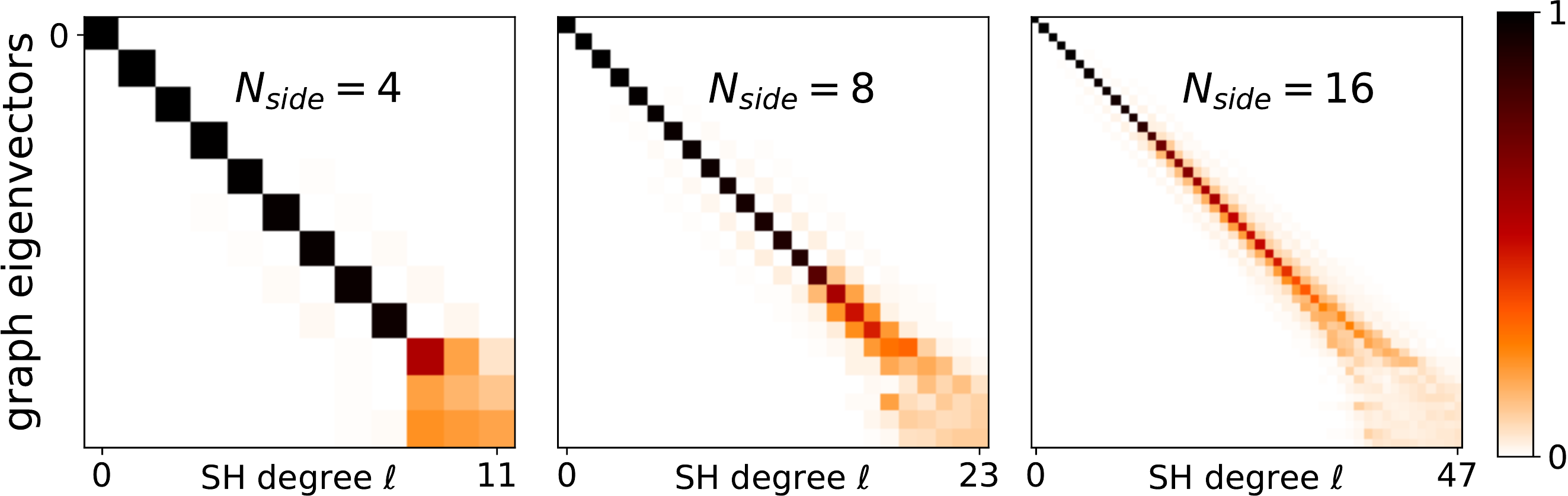}
	\caption{Correspondence between the subspaces spanned by the graph Fourier basis and the spherical harmonics.
		The eigenvectors $\U = [\b u_1, \dots, \b u_{N_{pix}}]$ of the graph Laplacian $\L = \U \bLambda \U\trans$, which form the Fourier basis, clearly resemble the spherical harmonics.
		The first 8 are shown on the left.
		To quantify the correspondence, we compute the power spectral density (PSD) of each eigenvector with the SHT.
		Second, as there is $2\ell+1$ spherical harmonics for each degree $\ell$, we sum the contributions of the corresponding $2\ell+1$ eigenvectors.
		The matrices on the right show how the subspaces align: the Fourier basis spans the same subspaces as the spherical harmonics in the low frequencies, and the eigenvectors leak towards adjacent frequency bands at higher frequencies.
	The graph Fourier basis aligns at higher frequencies as the resolution ($N_{pix} = 12 N_{side}^2$) increases.}
		\label{fig:subspace_harmonics_eigenvectors}
\end{figure}

Both the graph and the spherical convolutions can be expressed as multiplications in a Fourier domain.
As the spectral bases align, the two operations become similar.
Hence, the equivariance property of the spherical convolution carries out to the graph convolution.

\paragraph{Known case: the circle $\mathcal{C}^1$.}
Let $\theta\in[0,2\pi[$ be a parametrization of each point $(\cos\theta,\sin\theta)$ of $\mathcal{C}^1$.
The eigenfunctions of the Laplace-Beltrami operator of $\mathcal{C}^1$ are $u_\ell(\theta)=e^{i \theta m \ell}$, for $\ell \in \mathbb{N}$ and $m\in\{-1,1\}$.
Interestingly, for a \emph{regular sampling} of $\mathcal{C}^1$ (shown in \figref{ring}), the sampled eigenfunctions turn out to be the discrete Fourier basis.
That is, the harmonic decomposition of a discretized function on the circle can be done using the well-known discrete Fourier transform (DFT).
Moreover, the graph Laplacian of the sampled circle is diagonalized by the DFT basis, as all circulant matrices have complex exponentials as eigenbases \cite{strang1999dct}.
Hence, for $\mathcal{C}^1$, it can be verified, under mild assumptions, that the graph convolution is equivariant to translation \citep[section 2.2 and equation 3]{perraudin2017stationary}.
More generally, higher dimensional circles such as the torus $\mathcal{C}^2$ also have a circulant Laplacian and an equivariant convolution.
The above does however not hold for irregular samplings: the more irregular the sampling, the further apart the graph Fourier basis will be to the sampled eigenfunctions.

\paragraph{Analysis of the graph Laplacian used in DeepSphere.}
As there is no regular sampling of the sphere, we cannot have a similar reasoning as we had for the circle.
We can however perform a harmonic analysis to empirically assess how similar the graph and the spherical convolutions are.
The graph Laplacian eigenvalues, shown in \figref{graph_eigenvalues}, are clearly organized in frequency groups of $2\ell + 1$ orders for each degree $\ell$.
These blocks correspond to the different eigenspaces of the spherical harmonics.
We also show the correspondence between the subspaces spanned by the graph Fourier basis and the spherical harmonics in \figref{subspace_harmonics_eigenvectors}.
For example with $N_{side}=4$, we observe a good alignment for $\ell \leq 8$: the graph convolution will be equivariant to rotations for low frequency filters.
The imperfections are likely due to the small irregularities of the HEALPix sampling (varying number of neighbors and varying distances between vertices).
Furthermore, a follow-up study is underway to optimally construct the graph.
We also hope to get a proof of equivalence or convergence.



\begin{figure}[t!]
	\centering
	\includegraphics[width=0.32\linewidth]{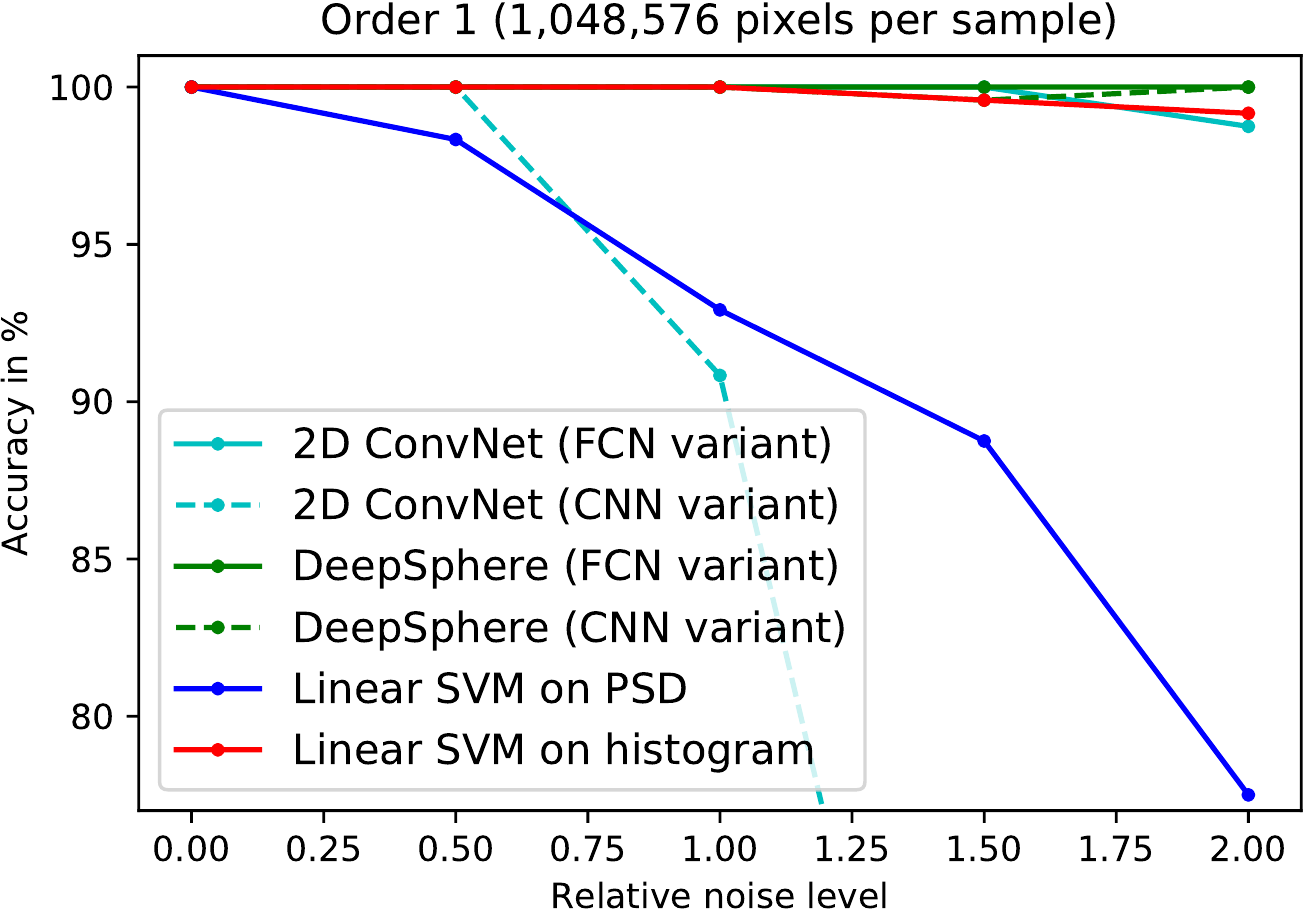}
	\hfill
	\includegraphics[width=0.32\linewidth]{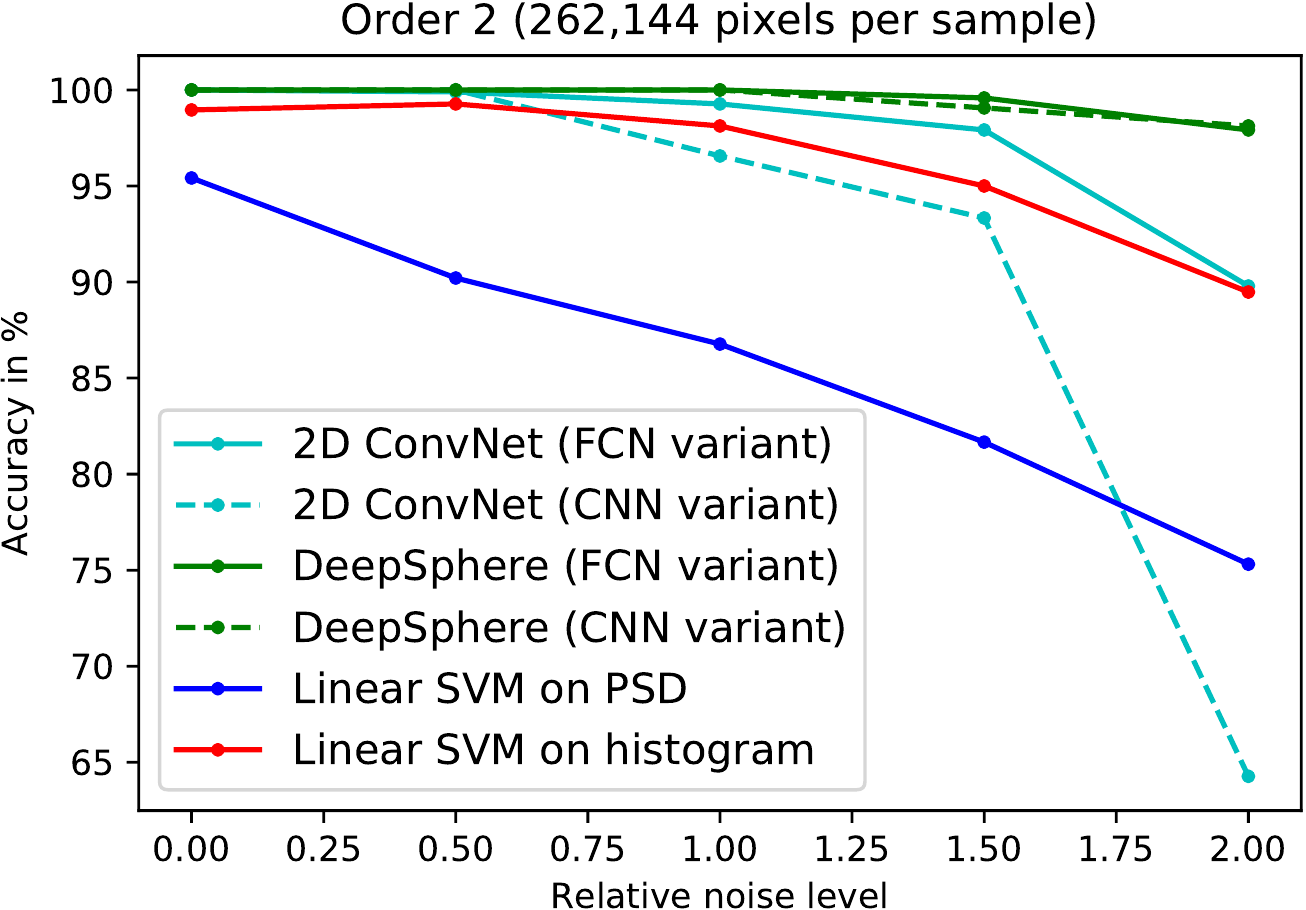}
	\hfill
	\includegraphics[width=0.32\linewidth]{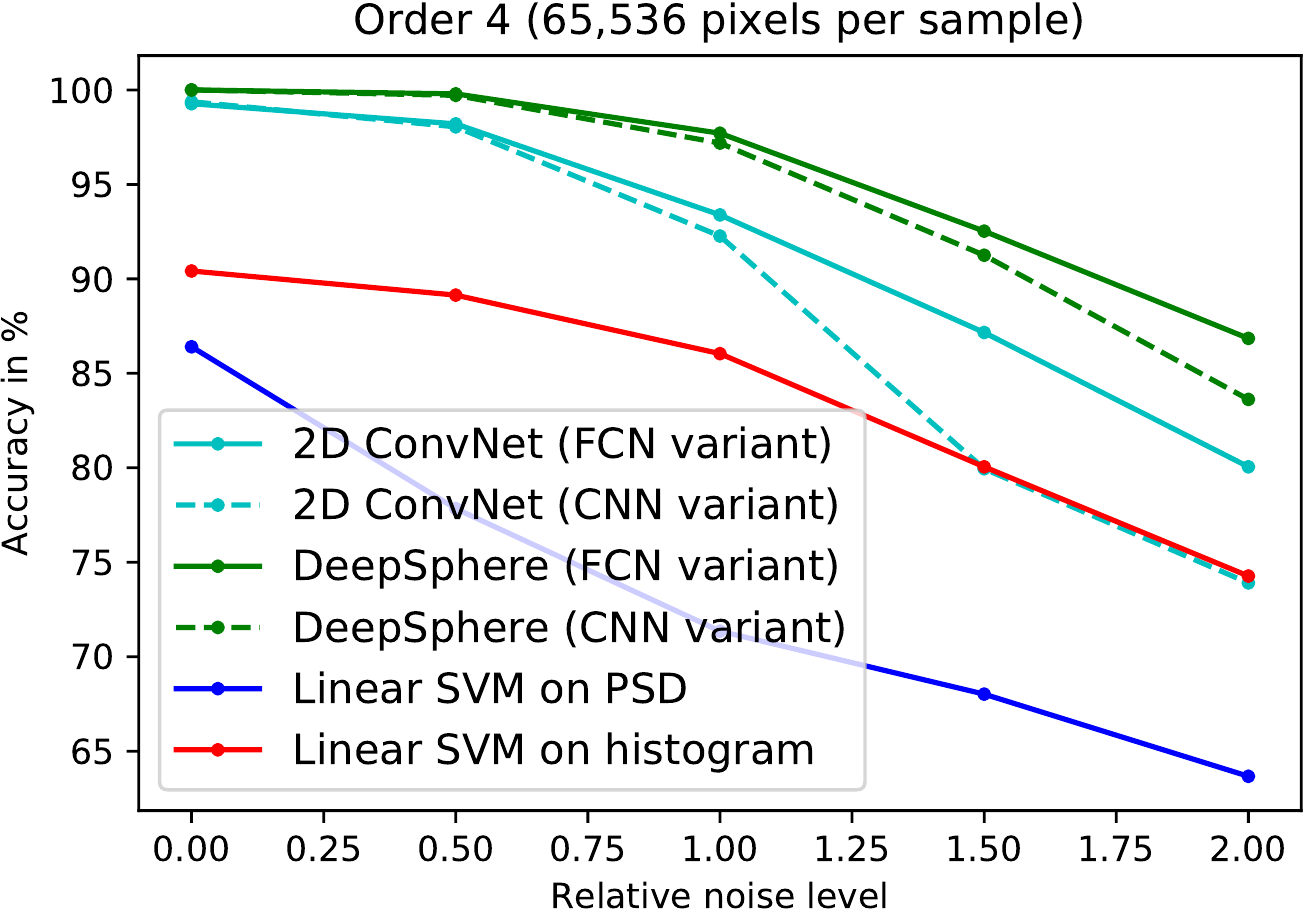}
	\caption{
		Classification accuracies.
		The difficulty of the task depends on the level of noise and the size of a sample (that is, the number of pixels that constitute the sample to classify). Order $o=1$ corresponds to samples which area is $\frac{1}{12}=8.1\%$ of the sphere ($\approx 1 \times 10^6$ pixels), order $o=2$ to $\frac{1}{12 \times 2^2} = 2.1\%$ ($\approx 260 \times 10^3$ pixels), and order $o=4$ to $\frac{1}{12 \times 4^2} = 0.5\%$ ($\approx 65 \times 10^3$ pixels).
		The FCN variant of DeepSphere beats the CNN variant by being invariant to rotation. Both variants largely beat the 2D ConvNet and the two SVM baselines.
	}
	\label{fig:results}
\end{figure}

\section{Experiments}


\paragraph{Setup.}
The performance of DeepSphere is demonstrated on a discrimination problem: the classification of cosmological convergence maps\footnote{Convergence maps represent the distribution of over- and under-densities of mass in the universe. They were created using the fast lightcone method UFalcon described in \citep{sgier2018fastgeneration}.} into two model classes.
Details about the experimental setup can be found in \cite{perraudin2018deepsphere}.
Code to reproduce the results is in the git repository.


We propose two architectures: the ``CNN variant'', where convolutional layers are followed by dense layers, and the ``FCN variant'', where convolutional layers are followed by global average pooling.
The FCN variant is rotation invariant to the extent that the convolution is rotation equivariant.

\paragraph{Baselines.}
DeepSphere is first compared against two simple yet powerful cosmological baselines, based on features that are (i) the power spectral densities (PSD) of maps, and (ii) the histogram of pixels in the maps \citep{patton2017cosmologicalconstraints}.
DeepSphere is further compared to a classical CNN for 2D Euclidean grids, as in \citep{krachmalnicoff2019convolutional}.
To be fed into the 2D ConvNet, partial spherical signals are transformed into flat images.
DeepSphere and the 2D ConvNet have the same number of trainable parameters.

\paragraph{Results.}
\Figref{results} summaries the results.
Overall, DeepSphere performs best with a gap that widens as the problem gets more difficult.
The FCN variant outperforms the CNN variant as it exploits the rotational symmetry of the task (explained by the cosmological principle of homogeneity and isotropy).
The 2D ConvNet fairs in-between the SVMs and DeepSphere.
Lower performance is thought to be due to a lack of rotation equivariance, and to the geometric distortion introduced by the projection, which the NN has to learn to compensate for.

\subsubsection*{Acknowledgments}

We thank Pierre Vandergheynst for advice and helpful discussions, and Andreas Loukas for having processed the raw GHCN data.
The Python Graph Signal Processing package (PyGSP) \citep{pygsp} was used to build graphs, compute the Laplacian and Fourier basis, and perform graph convolutions.

\bibliography{refs}
\bibliographystyle{iclr2019}

\end{document}